\documentclass{article}

\usepackage{PRIMEarxiv}

\usepackage[utf8]{inputenc} 
\usepackage[T1]{fontenc}    
\usepackage{hyperref}       
\usepackage{url}            
\usepackage{booktabs}       
\usepackage{amsfonts}       
\usepackage{nicefrac}       
\usepackage{microtype}      
\usepackage{lipsum}
\usepackage{fancyhdr}       
\usepackage{graphicx}       
\graphicspath{{media/}}     

\title{Prevalent Frequency of Emotional and Physical Symptoms in Social
Anxiety using Zero Shot Classification: An Observational Study
\thanks{The article is published in the proceedings of The Workshop on Computational Linguistics and Clinical Psychology (CLPsych) 2024.} 
}

\author{
  Muhammad Rizwan \\
  Department of Information Technology, \\
  University of Engineering\\ and Information Technology, \\
  Rahim Yar Khan, Pakistan.\\
  \texttt{rizwan2phd@gmail.com} \\
   \And
  Jure Demšar \\
  Faculty of Computer and \\
  Information Studies, \\
  University of Ljabljana, \\
  Slovenia.\\
  \texttt{Jure.Demsar@fri.uni-lj.si} \\
}

\begin{document}
\maketitle

\begin{abstract}
Social anxiety represents a prevalent challenge in modern society, affecting individuals across personal and professional spheres. Left unaddressed, this condition can yield substantial negative consequences, impacting social interactions and performance. Further understanding its diverse physical and emotional symptoms becomes pivotal for comprehensive diagnosis and tailored therapeutic interventions. This study analyze prevalence and frequency of social anxiety symptoms taken from Mayo Clinic, exploring diverse human experiences from utilizing a large Reddit dataset dedicated to this issue. Leveraging these platforms, the research aims to extract insights and examine a spectrum of physical and emotional symptoms linked to social anxiety disorder. Upholding ethical considerations, the study maintains strict user anonymity within the dataset. By employing a novel approach, the research utilizes BART-based multi-label zero-shot classification to identify and measure symptom prevalence and significance in the form of probability score for each symptom under consideration. Results uncover distinctive patterns: "Trembling" emerges as a prevalent physical symptom, while emotional symptoms like "Fear of being judged negatively" exhibit high frequencies. These findings offer insights into the multifaceted nature of social anxiety, aiding clinical practices and interventions tailored to its diverse expressions.
\end{abstract}


\section{Introduction}
Social anxiety is prevalent in our society, posing a significant and widespread difficulty for individuals. The impact is deep and goes beyond limits, affecting both personal and professional aspects. If not addressed, this illness can have a significant impact, leading to a series of negative consequences. Going beyond just being shy, its enduring nature can result in significant repercussions \cite{hur2020social, lepine2000take}, spanning from limited social contacts to compromised performance in several areas of life. The consistent existence of this phenomenon in the lives of many emphasises the importance of understanding its complexities and identifying its various forms in order to provide appropriate intervention and assistance. Social anxiety is a mental health problem that can have long-term consequences \cite{blood2016long}. The long-lasting character of the phenomenon emphasises the crucial importance of thoroughly analysing its physical and emotional symptoms, in order to fully comprehend its frequency and influence on the individuals afflicted \cite{liu2023behavioral, zech2023safety}. Understanding these symptoms not only helps to make the diagnosis clear but also facilitates the development of customised therapies, enabling prompt and accurate assistance for individuals struggling with this incapacitating condition.

In order to gain a comprehensive understanding of this complex subject, this study undertakes an investigative exploration into the realm of symptoms associated with social anxiety. This study aims to extract insights by analysing the diverse range of human experiences reported on Reddit subreddits dedicated to this issue. These platforms provide an unedited view of the real-life experiences of people dealing with the intricacies of social anxiety, including a large amount of text data that is suitable for research. By utilising this highly important resource, the research seeks to examine and define the intricate range of physical and emotional symptoms linked to social anxiety disorder. We strictly ensured that ethical and privacy consideration during our analysis, does not reveal any reddit user identity in the study. 

This work utilises a new method by applying a multi-class zero-shot classification strategy assisted by BART (Bidirectional and Auto-Regressive Transformers). This framework utilises NLP and deep learning to identify and measure common symptoms, leading to a better understanding of the complex nature of social anxiety with respect to different emotional and physical symptoms. BART zero shot classification has already been used for such studies e.g. in \cite{farruque2021explainable, yang2023mentalllama}. This research aims to provide which symptoms are more frequent than the others comparatively using deep  context provided by the BART language model. This study can further be used to inform clinical practices, interventions, and support mechanisms specifically designed to address the various ways in which social anxiety is expressed.

\begin{figure*}[hbt!]
  \includegraphics[width=\textwidth]{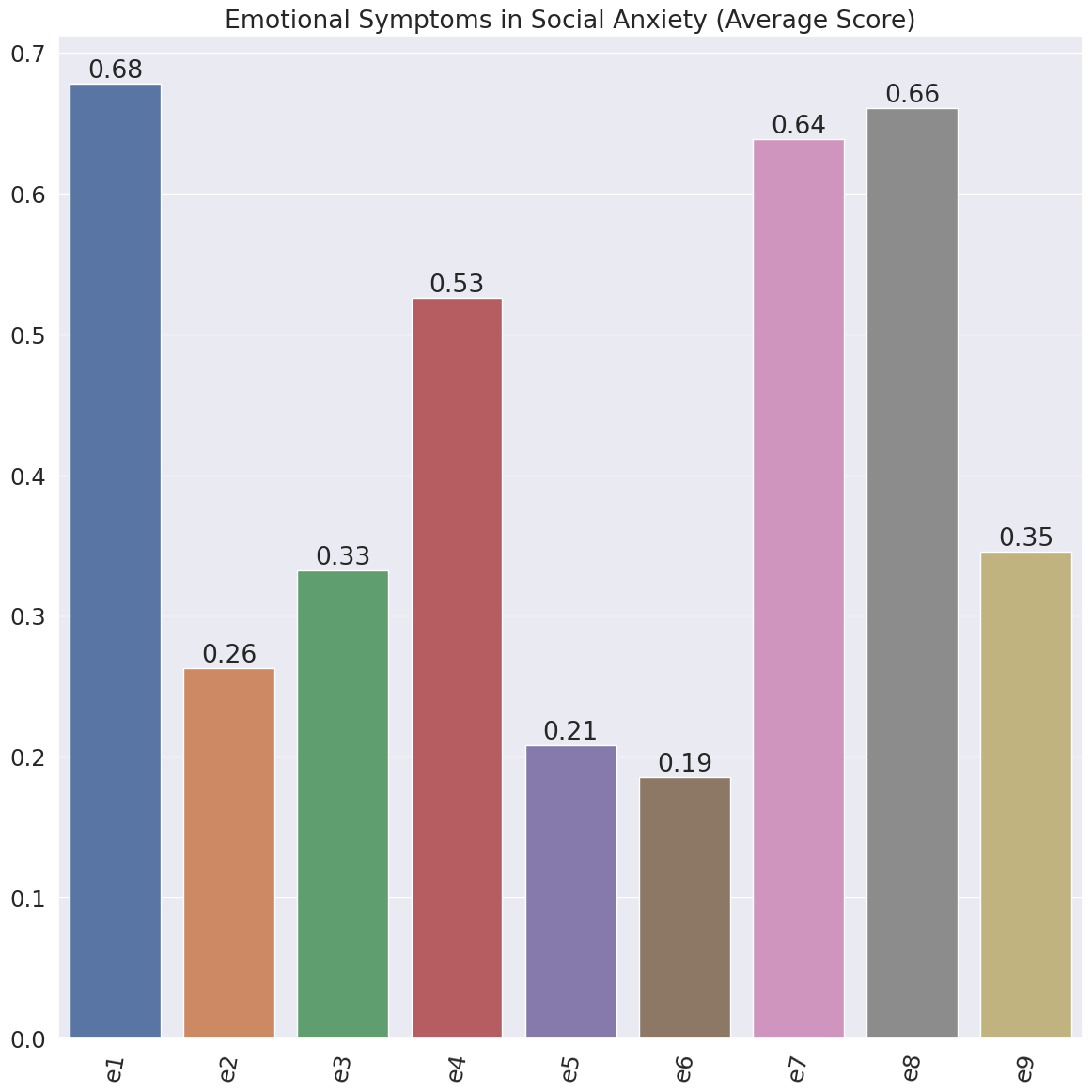}
  \caption{The bar chart illustrates the average zero-shot classification probability scores for emotional symptoms mentioned in Table 1 related to social anxiety. The scores were computed by averaging all individual scores for all 12,277 subreddits  text documents / subreddits.}
  \label{schizophrenia}
\end{figure*}
\begin{table*}[hbt!]
\centering
\begin{tabular}{p{0.7in}p{5.3in}}
\hline
\textbf{No.} & \textbf{Emotional Symptoms} \\
\hline
e1&Fear of situations in which you may be judged negatively  \\
e2&Worry about embarrassing or humiliating yourself \\
e3&Intense fear of interacting or talking with strangers \\
e4&Fear that others will notice that you look anxious \\
e5&Avoidance of doing things or speaking to people out of fear of embarrassment \\
e6&Avoidance of situations where you might be the center of attention \\
e7&Anxiety in anticipation of a feared activity or event \\
e8&Intense fear or anxiety during social situations \\
e9&Expectation of the worst possible consequences from a negative experience during a social situation \\
\hline
\end{tabular}
\caption{\label{citation-guide}
The table illustrates the emotional symptoms associated with social anxiety as outlined on the Mayo Clinic \cite{Mayo_Clinic_2021} 
}
\end{table*}

\section{Social Anxiety Symptoms}
This study considers the following common symptoms facing by common people during social anxiety disorder in order to analysis their prevalence frequency in reddit social anxiety dataset.
\subsection{Physical Symptoms}
We selected common physical symptoms associated with social anxiety disorder, as listed on the Mayo Clinic website \cite{Mayo_Clinic_2021} mentioned in Table 1 and 2, for deeper analysis within the Reddit dataset. These symptoms encompass a range of experiences, including blushing, a rapid heartbeat, trembling, sweating, upset stomach or nausea, difficulty breathing, feelings of dizziness or lightheadedness, experiencing mental blankness, and muscle tension. These specific manifestations represent key indicators of the physical impact that social anxiety can exert on individuals, prompting our exploration within the Reddit dataset to glean insights and understand the prevalence frequency of these symptoms in real life.

\subsection{Emotional Symptoms}
We have also extracted common emotional symptoms associated with social anxiety disorder from the Mayo Clinic website. These symptoms encompass a range of experiences, including a pervasive fear of negative judgment in social situations, concerns about potential embarrassment or humiliation, and an intense fear of interacting with strangers. Additionally, these symptoms encompass a fear of others noticing anxious behavior, avoidance of social interactions due to fear of embarrassment, and evading situations where attention might be directed toward oneself. The anticipation of anxiety-inducing activities, intense fear or anxiety during social interactions, and expecting the worst possible outcomes from negative experiences within social settings also form part of these emotional symptoms. These descriptors serve as crucial elements for further analysis within the Reddit dataset, providing a comprehensive understanding of the emotional complexities experienced by individuals grappling with social anxiety disorder.

\begin{table}[hbt!]
\centering
\begin{tabular}{p{0.4in}p{2.4in}}
\hline
\textbf{No.} & \textbf{Physical Symptoms} \\
\hline
p1& Blushing  \\
p2&Fast heartbeat  \\
p3&Trembling \\
p4&Sweating \\
p5&Upset stomach or nausea \\
p6&Trouble catching your breath \\
p7&Dizziness or lightheadedness \\
p8&Feeling that your mind has gone blank \\
p9&Muscle tension \\
\hline
\end{tabular}
\caption{\label{citation-guide}
The table illustrates the physical symptoms associated with social anxiety as outlined on the Mayo Clinic \cite{Mayo_Clinic_2021} }
\end{table}
\begin{figure*}[hbt!]
  \includegraphics[width=\textwidth]{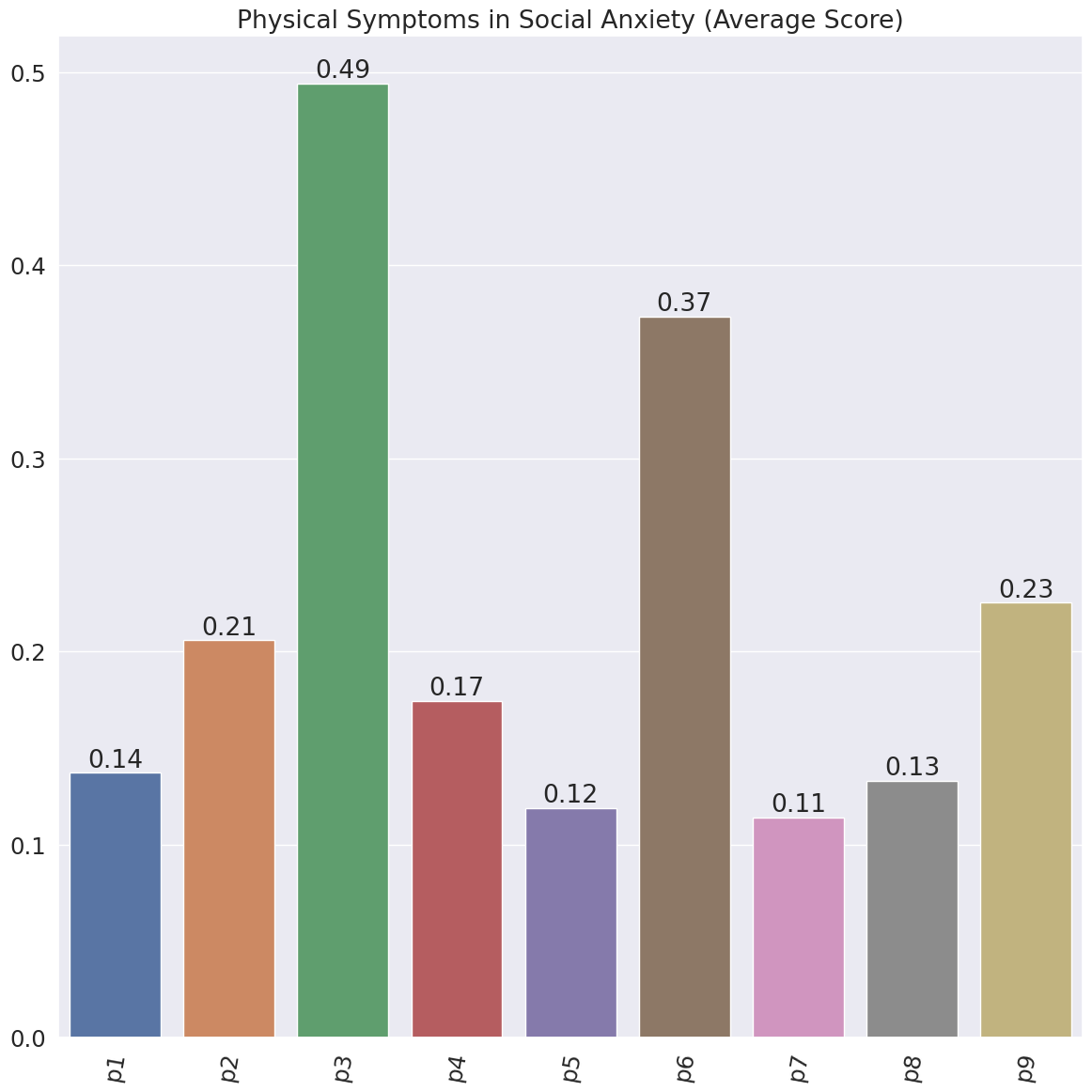}
  \caption{The bar chart illustrates the average zero-shot classification probability scores for physical symptoms mentioned in Table 2 related to social anxiety. The scores were computed by averaging all individual scores for all 12,277 subreddits text documents / subreddits.}
  \label{schizophrenia}
\end{figure*}

\section{Method}
In this methodology section, we'll first outline the Reddit dataset chosen for the study on social anxiety disorder. This dataset forms the core of our investigation. Next, we introduce and examine the emotional and physical symptoms associated with social anxiety disorder, employing a zero-shot classification approach. This method allows us to explore a wide array of symptoms without requiring specific training data. Finally, we provide a detailed explanation of BART (Bidirectional and Auto-Regressive Transformers) and its utilization in a multi-label zero-shot classification setup. This methodology enables us to calculate the average probability of each social anxiety symptom within the dataset, offering a comprehensive insight into their prevalence and significance within the scope of this research.

\subsection{Social Anxiety Reddit Dataset}
The dataset utilized in this research, as detailed by \cite{low2020natural}, was acquired using the pushshift API of Reddit. Researchers gathered posts from 15 distinct subreddits dedicated to various mental health communities. These subreddits encompassed a wide range of mental health concerns, including communities like r/EDAnonymous, r/addiction, r/alcoholism, r/adhd, r/anxiety, r/autism, r/BipolarReddit, r/bpd, r/depression, r/healthanxiety, r/lonely, r/ptsd, r/schizophrenia, r/socialanxiety, and r/SuicideWatch.

In the context of our study's specific objectives, we narrow our focus to the subreddit r/socialanxiety, honing in on discussions related to social anxiety disorder. The dataset under consideration comprises 12,277 text documents or subreddits, capturing social anxiety-related content posted between 2018 and 2019. Within the dataset of the r/socialanxiety Reddit community, individuals engage in discussions about their real-life experiences, opinions, and symptoms related to social anxiety. Notably, each document is associated with a distinct user, resulting in a dataset intentionally diversified with contributions from 12,277 unique users—a deliberate choice to enhance reliability in the context of crowd-sourcing tasks.

Our purpose in utilizing this dataset is to delve into the unique perspectives and challenges voiced by individuals within these online communities. The central objective is to conduct a thorough analysis, aiming to comprehend the intricate interplay and associations among various prevalent symptoms discussed within the r/socialanxiety subreddit. Our exploration involves examining the diverse experiences shared within this particular online community, with the primary goal of uncovering and scrutinizing the complex relationships between different symptoms linked to social anxiety disorder. This investigative approach provides a distinctive opportunity to gain valuable insights into the multifaceted nature of social anxiety, as perceived and expressed by members of this specific online community.

\subsection{BART Based Multi-Label Zero Shot Classification}
Facebook AI Research (FAIR) is accountable for the development of BART (Bidirectional and Auto Regressive Transformer), a progressive language model \cite{lewis2019bart}. The model is pretrained using a combination of denoising autoencoding and sequence-to-sequence tasks, and it is based on the Transformer architecture. The architecture of the BART model consists of encoders and decoders. The encoder receives the input sequence and proceeds to process it through a sequence of transformer layers. Every transformer layer includes position-wise feed-forward neural networks alongside multi-head self-attention approaches. The model can effectively capture the relationships between individual words in the input sequence through a technique known as self-attention.

\begin{figure*}[hbt!]
  \includegraphics[width=\textwidth]{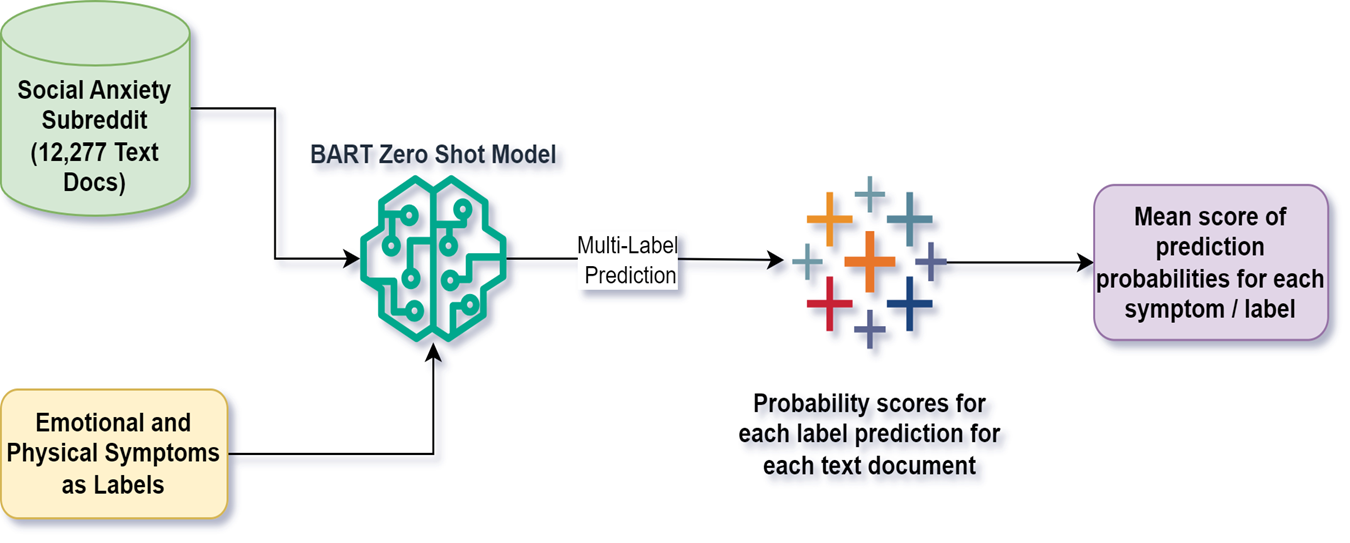}
  \caption{The workflow of the study.}
  \label{workf}
\end{figure*}

The encoder generates an encoded representation, which is subsequently handed to the decoder. The decoder then produces the output sequence in an autoregressive manner. In addition to utilising transformer layers, it also has a cross-attention mechanism that focuses on the encoded input sequence. As a result, the model has the capability to produce output tokens that depend not just on the input sequence but also on the tokens it has previously generated. During the pre-training phase, BART undergoes training using vast amounts of data, which can be either monolingual or parallel. It gains the capacity to reconstruct the initial sequence of input data from damaged copies, allowing it to better capture significant representations of the entered data. The versatility of BART in performing various text generation tasks, including text summarization, machine translation, and text completion, is a very notable feature of this application. By conducting fine-tuning on specific downstream tasks and adjusting the model accordingly, it is feasible to optimise the pre-trained BART model to generate high-quality outputs for a diverse range of natural language processing applications.

\cite{yin2019benchmarking} introduced an innovative technique harnessing the capabilities of pre-trained Natural Language Inference (NLI) models as adept zero-shot sequence classifiers. \cite{tesfagergish2022zero, chae2023large} This approach involves structuring the sequence under examination as the NLI premise, then formulating a hypothesis for each potential label. For instance, when scrutinizing whether a sequence aligns with a specific social anxiety symptom, such as "Trembling," a corresponding hypothesis might read, "This text is about Trembling."  Following this framing, the probabilities associated with entailment (alignment) and contradiction (misalignment) undergo transformation into probabilities specifically linked to each symptom label \cite{patadia2021zero, basile2021probabilistic}. In instances where multiple labels could be pertinent, activating the multi-label setting, utilizing the huggingface BART implementation, enables the independent computation of probabilities for each symptom class.

The BART-based zero-shot classification solutions distinguish themselves from traditional supervised learning classification approaches by leveraging pre-trained language models. In contrast to supervised learning, which necessitates labeled training data for each class, BART-based zero-shot classification can generalize to new, unseen classes without specific training on them \cite{moreno2023novel}. This adaptability renders BART-based solutions superior to traditional classification algorithms. It proves particularly advantageous when dealing with evolving or dynamic datasets where labeled examples for all potential classes may not be readily accessible.

In our study, we employ this methodology to calculate the probabilities for each label representing various social anxiety symptoms within each individual Reddit text document. This meticulous analysis allows us to determine the relative strength of association between the document and each symptom label. Subsequently, by computing the arithmetic mean of these probabilities across all documents for each symptom label, we unveil the prevalence frequency of each social anxiety symptom within the dataset. This rigorous process provides a comprehensive understanding of the varying degrees of manifestation for different symptoms within the context of social anxiety expressed in Reddit conversations. The workflow of the study can be seen in Figure \ref{workf}.

\section{Results and Discussion}
Tables 1 and 2 present a comprehensive list of both physical and emotional symptoms of social anxiety utilized in this study. Additionally, Figures 1 and 2 display bar charts illustrating the average zero-shot classification probability scores for these symptoms. Figure 1 focuses on the physical symptoms outlined in Table 2, showcasing the average zero-shot classification probability scores related to social anxiety. These scores were calculated by averaging individual scores across all 12,277 subreddit text documents. Similarly, Figure 2 depicts the average zero-shot classification probability scores for emotional symptoms from Table 1 associated with social anxiety. The computation involved averaging individual scores across the entire set of 12,277 subreddit text documents.

Beginning with an analysis of the results concerning physical symptoms, the data reveals intriguing patterns. Notably, the most prevalent physical symptom observed within social anxiety disorder emerges as "Trembling," boasting significantly, average probability score of 49 percent. Trembling, an outward manifestation characterized by body shaking, stands out prominently within this context, reflecting its substantial association with social anxiety experiences.

Following closely after Trembling, the subsequent prominent physical symptom is the sensation of catching one's breath. This finding reinforces the study's validity by mirroring the reality experienced by numerous individuals worldwide grappling with social anxiety. Moreover, the data highlights additional moderately prevalent symptoms, including rapid heartbeat, sweating, and muscle tension, each scoring approximately around 20 percent. This categorization positions Trembling and catching breath as higher-frequency symptoms, while the trio of rapid heartbeat, sweating, and muscle tension forms the middle tier in terms of prevalence.  Conversely, the analysis also unveils relatively lower-frequency physical symptoms of social anxiety. Symptoms such as blushing, mental blankness, and upset stomach emerge with probability scores ranging from approximately 11 to 14 percent. These findings collectively delineate a gradient of symptom prevalence within the spectrum of social anxiety, showcasing the varying degrees of manifestation experienced by individuals grappling with this condition \cite{heerey2007interpersonal, weeks2008fear}.

When examining the prevalence and frequency of emotional symptoms, notable trends surface within the dataset. Particularly striking is the prominence of three emotional symptoms, each registering a notably high frequency: "Fear of being judged negatively," "Anxiety or fear of events," and "Intense fear of social situations." These three emotions exhibit probability scores ranging from 64 to 68 percent, signifying their substantial occurrence among individuals grappling with social anxiety disorder. Additionally, the analysis uncovers that the fear of others noticing one's nervousness ranks as the second-highest emotional concern among respondents. Following closely behind, in the third position, are two emotional states: "Fear of worst consequences due to negative social experiences" and "Fear of talking with strangers," each carrying a notable probability score of 33 to 35 percent.

Conversely, the emotional symptoms exhibiting the lowest probability scores are "Avoidance of situations where attention might be drawn" and "Avoidance of speaking to people due to fear of embarrassment." These findings offer nuanced insights into the varying degrees of emotional distress experienced by individuals dealing with social anxiety disorder. These intriguing discoveries \cite{beard2008multi}, obtained through observational studies, hold significant potential for fostering a deeper understanding of social anxiety from a community perspective. Moreover, they pave the way for crowd-sourced insights, potentially contributing to the development of innovative remedies and interventions aimed at alleviating the challenges posed by social anxiety. The insights gleaned from these emotional symptom prevalence patterns can serve as a valuable resource in guiding future research and therapeutic strategies aimed at addressing this prevalent mental health issue.

As previously stated, the findings presented herein are derived from a comprehensive analysis of 12,277 subreddits belongs to social anxiety, each associated with distinct users. It is crucial to acknowledge the potential for bias in the information gathered from these diverse sources. However, to mitigate this concern, we have adopted a methodological approach that involves averaging the prevalent scores. This calculation entails determining the probability score for each symptom across all subreddits. In doing so, we strive to enhance the authenticity of the information, particularly in the context of crowd-sourced data.

Nevertheless, to further bolster the credibility and reliability of these observational results, it is imperative to seek validation through clinical verification. The integration of clinical assessments would provide a more robust foundation for the findings, ensuring a comprehensive and well-rounded evaluation of the presented information.

\section{Limitations}
This study explores the experiences of individuals coping with social anxiety disorder, utilising user-generated content from Reddit. Nevertheless, it is crucial to recognise the inherent limitations associated with this observational methodology. The collected data includes self-reported experiences obtained from an open platform, which may introduce biases in terms of accuracy and completeness.  Moreover, the absence of clinical validation or expert assessment of these symptoms could affect the precision and clinical significance of the collected information. This study solely captures experiences that are unique to the Reddit platform, perhaps disregarding a wide range of opinions or individuals, which in future may be verified from the clinical point of view for the further validation of this study.

\section*{Ethics Statement}
This study upholds stringent ethical and privacy considerations throughout its entirety. Specifically, the dataset sourced from Reddit is rigorously maintained under the Public Domain Dedication and License v1.0, ensuring the preservation of Reddit users' privacy. Importantly, the study maintains a strict adherence to anonymity, refraining from disclosing any user identities within the article or its findings. This commitment to confidentiality and privacy safeguards the individuals contributing to the dataset, upholding their anonymity and confidentiality in line with ethical standards.

\section*{Acknowledgments}
This was was supported in part by......

\bibliographystyle{unsrt}  
\bibliography{references}

\begin{thebibliography}{10}

\bibitem{hur2020social}
Juyoen Hur, Kathryn~A DeYoung, Samiha Islam, Allegra~S Anderson, Matthew~G Barstead, and Alexander~J Shackman.
\newblock Social context and the real-world consequences of social anxiety.
\newblock {\em Psychological Medicine}, 50(12):1989--2000, 2020.

\bibitem{lepine2000take}
Jean-Pierre L{\'e}pine and Antoine Pelissolo.
\newblock Why take social anxiety disorder seriously?
\newblock {\em Depression and Anxiety}, 11(3):87--92, 2000.

\bibitem{blood2016long}
Gordon~W Blood and Ingrid~M Blood.
\newblock Long-term consequences of childhood bullying in adults who stutter: Social anxiety, fear of negative evaluation, self-esteem, and satisfaction with life.
\newblock {\em Journal of fluency disorders}, 50:72--84, 2016.

\bibitem{liu2023behavioral}
Pan Liu and Jaron~XY Tan.
\newblock Behavioral and erp indices of self-schematic processing show differential associations with emerging symptoms of depression and social anxiety in late childhood: Evidence from a community-dwelling sample.
\newblock {\em Biological Psychology}, page 108594, 2023.

\bibitem{zech2023safety}
James~M Zech, Tapan~A Patel, and Jesse~R Cougle.
\newblock Safety behaviors predict long-term treatment outcome following internet-based treatment of adults with social anxiety disorder.
\newblock {\em Cognitive Therapy and Research}, 47(3):412--422, 2023.

\bibitem{farruque2021explainable}
Nawshad Farruque, Randy Goebel, Osmar~R Za{\"\i}ane, and Sudhakar Sivapalan.
\newblock Explainable zero-shot modelling of clinical depression symptoms from text.
\newblock In {\em 2021 20th IEEE International Conference on Machine Learning and Applications (ICMLA)}, pages 1472--1477. IEEE, 2021.

\bibitem{yang2023mentalllama}
Kailai Yang, Tianlin Zhang, Ziyan Kuang, Qianqian Xie, and Sophia Ananiadou.
\newblock Mentalllama: Interpretable mental health analysis on social media with large language models.
\newblock {\em arXiv preprint arXiv:2309.13567}, 2023.

\bibitem{Mayo_Clinic_2021}
Jun 2021.

\bibitem{low2020natural}
Daniel~M Low, Laurie Rumker, Tanya Talkar, John Torous, Guillermo Cecchi, and Satrajit~S Ghosh.
\newblock Natural language processing reveals vulnerable mental health support groups and heightened health anxiety on reddit during covid-19: Observational study.
\newblock {\em Journal of medical Internet research}, 22(10):e22635, 2020.

\bibitem{lewis2019bart}
Mike Lewis, Yinhan Liu, Naman Goyal, Marjan Ghazvininejad, Abdelrahman Mohamed, Omer Levy, Ves Stoyanov, and Luke Zettlemoyer.
\newblock Bart: Denoising sequence-to-sequence pre-training for natural language generation, translation, and comprehension.
\newblock {\em arXiv preprint arXiv:1910.13461}, 2019.

\bibitem{yin2019benchmarking}
Wenpeng Yin, Jamaal Hay, and Dan Roth.
\newblock Benchmarking zero-shot text classification: Datasets, evaluation and entailment approach.
\newblock {\em arXiv preprint arXiv:1909.00161}, 2019.

\bibitem{tesfagergish2022zero}
Senait~Gebremichael Tesfagergish, Jurgita Kapo{\v{c}}i{\=u}t{\.e}-Dzikien{\.e}, and Robertas Dama{\v{s}}evi{\v{c}}ius.
\newblock Zero-shot emotion detection for semi-supervised sentiment analysis using sentence transformers and ensemble learning.
\newblock {\em Applied Sciences}, 12(17):8662, 2022.

\bibitem{chae2023large}
Youngjin Chae and Thomas Davidson.
\newblock Large language models for text classification: From zero-shot learning to fine-tuning.
\newblock {\em Open Science Foundation}, 2023.

\bibitem{patadia2021zero}
Devika Patadia, Shivam Kejriwal, Pashva Mehta, and Abhijit~R Joshi.
\newblock Zero-shot approach for news and scholarly article classification.
\newblock In {\em 2021 International Conference on Advances in Computing, Communication, and Control (ICAC3)}, pages 1--5. IEEE, 2021.

\bibitem{basile2021probabilistic}
Angelo Basile, Guillermo P{\'e}rez-Torr{\'o}, and Marc Franco-Salvador.
\newblock Probabilistic ensembles of zero-and few-shot learning models for emotion classification.
\newblock In {\em Proceedings of the International Conference on Recent Advances in Natural Language Processing (RANLP 2021)}, pages 128--137, 2021.

\bibitem{moreno2023novel}
Carlos~Francisco Moreno-Garcia, Chrisina Jayne, Eyad Elyan, and Magaly Aceves-Martins.
\newblock A novel application of machine learning and zero-shot classification methods for automated abstract screening in systematic reviews.
\newblock {\em Decision Analytics Journal}, page 100162, 2023.

\bibitem{heerey2007interpersonal}
Erin~A Heerey and Ann~M Kring.
\newblock Interpersonal consequences of social anxiety.
\newblock {\em Journal of abnormal psychology}, 116(1):125, 2007.

\bibitem{weeks2008fear}
Justin~W Weeks, Richard~G Heimberg, and Thomas~L Rodebaugh.
\newblock The fear of positive evaluation scale: Assessing a proposed cognitive component of social anxiety.
\newblock {\em Journal of anxiety disorders}, 22(1):44--55, 2008.

\bibitem{beard2008multi}
Courtney Beard and Nader Amir.
\newblock A multi-session interpretation modification program: Changes in interpretation and social anxiety symptoms.
\newblock {\em Behaviour research and therapy}, 46(10):1135--1141, 2008.

\end{thebibliography}

\end{document}